\documentclass[10pt, journal]{IEEEtran}
\usepackage{graphics}
\usepackage{graphicx}
\usepackage{epstopdf}
\usepackage{amsmath}
\usepackage{amssymb}
\usepackage{url}
\usepackage[usenames,dvipsnames]{color}
\usepackage{dsfont}
\usepackage{enumerate}
\usepackage{cite}
\usepackage{mymacros}
\newtheorem{theorem}{Theorem}
\newtheorem{lemma}{Lemma}

\newtheorem{cor}{Corollary}
\newtheorem{assumption}{Assumption}

\DeclareMathAlphabet{\mathpzc}{OT1}{pzc}{m}{it}

\ifCLASSINFOpdf
\else
\fi
\begin{document}
%
\title{Geometric Learning and Topological Inference with Biobotic Networks: Convergence  Analysis}
%
%
%

\author{Alireza~Dirafzoon$^1$,
	Alper~Bozkurt$^1$,
	and Edgar~Lobaton$^1$
    \thanks{This work was supported by the National Science Foundation under award CNS-1239243.}
	\thanks{$^1$A. Dirafzoon, A. Bozkurt, and E. Lobaton   are with the Department
		of Electrical and Computer Engineering, North Carolina State University, Raleigh,
		NC, 27606 USA. e-mail: adirafz@ncsu.edu, aybozkur@ncsu.edu, ejlobato@ncsu.edu. }
	}

\maketitle



%
\IEEEpeerreviewmaketitle
\begin{abstract}


In this study, we present and analyze a framework for geometric and topological estimation for mapping of unknown environments.
We consider agents mimicking   
motion behaviors of cyborg insects, known as \textit{biobots}, and exploit  coordinate-free local interactions among them to infer geometric and topological information about the environment,
under minimal sensing and localization constraints. Local interactions are used to create a graphical representation referred  to as the encounter graph.
A metric is estimated over the encounter graph of the agents in order to construct a geometric point cloud using manifold learning techniques.  
Topological data analysis (TDA), in particular persistent homology, is used 
in order to extract topological features of the space and a classification method is proposed to infer robust features of interest (e.g. existence of obstacles).  
We examine the asymptotic behavior of the proposed metric in terms of the convergence to the geodesic distances in the underlying manifold of the domain, and provide stability analysis results for the topological persistence.  
The proposed framework and its convergences and stability analysis are demonstrated through numerical simulations and experiments.


\end{abstract}
\section{Introduction} \label{sec:intro}

\begin{figure}[tb]
      \centering
   \includegraphics[width=0.9\linewidth]{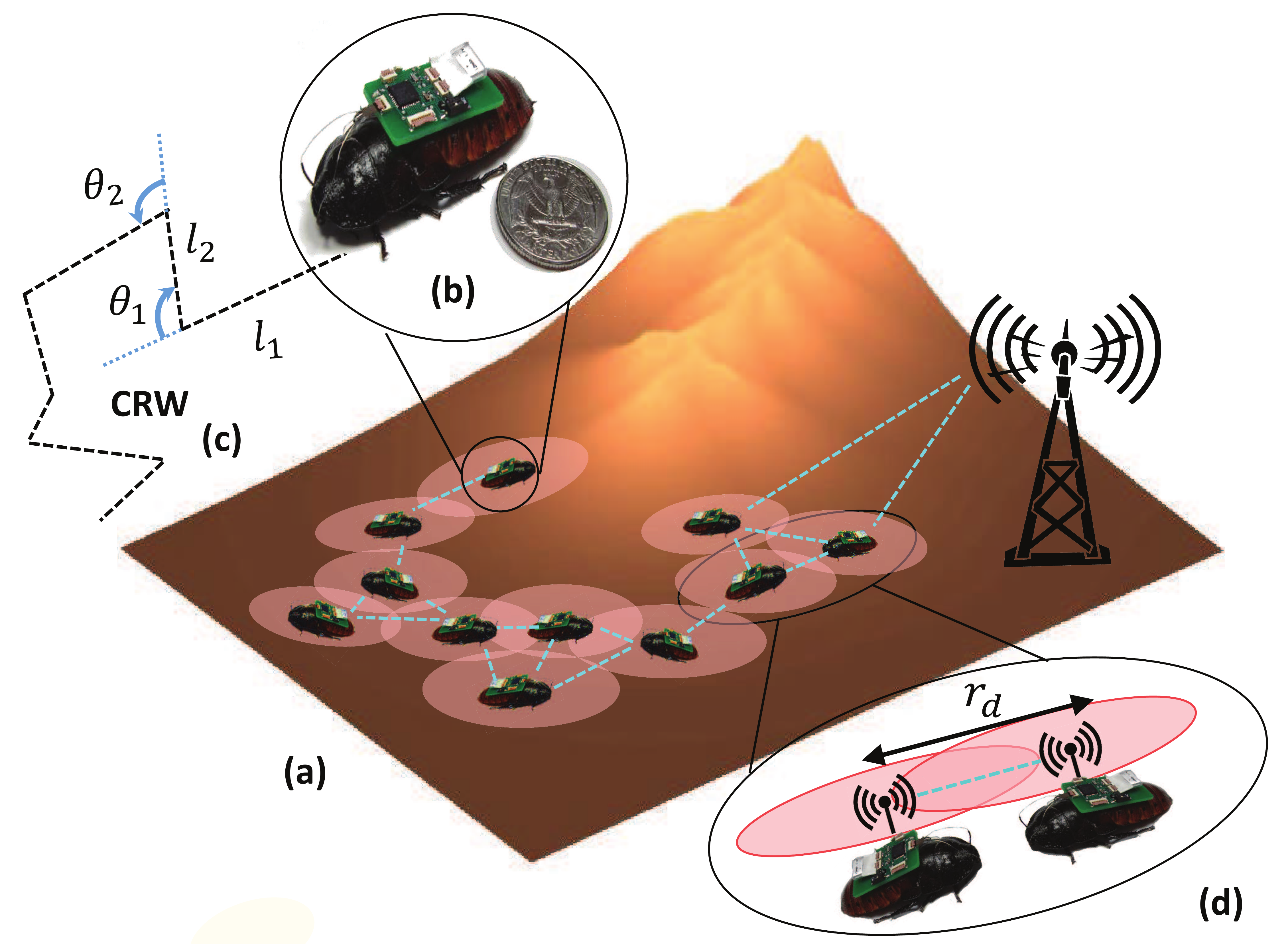}
      \caption{ \small {(a) A physical environment with a biobotic (cyborg insect)
      network, (b) \textit{Gromphadorhina portentosa} ``biobot'' with stimulation backpack and battery \cite{latif2012}, (c) natural motion model for the biobots  in terms of correlated random walk (CRW), (d) encounter region for two biobots.}
 }
      \label{fig:front}
   \end{figure}

\IEEEPARstart{R}{ecent }  
 developments in neural engineering have empowered us to directly control insect locomotion using wireless neuro-stimulators to enable remotely controlled cyborg insects, known as \textit{insect biobots} \cite{latif2012}. 
 Each agent is equipped with a system-on-chip based ZigBee enabled wireless neuro-stimulation backpack system and remote navigational control circuits. In particular, we have been able to control \textit{Gromphadorhina portentosa} (Madagascar hissing) cockroaches in open and mazed environments
 \cite{whitmire2013kinect, latif2014towards, latif2012}.
  Exploiting unmatched natural abilities of insects in navigation and exploration, a cyber-physical network of such biobots, which we call a \textit{biobotic network}, could prove useful for search and rescue applications in uncertain disaster environments (e.g. when an earthquake occurs).  Figure~\ref{fig:front} illustrates a realization of  a biobotic network, a snapshot of one of our cockroach biobots, and a summary of their natural mobility and sensing model. 

   

Exploration and mapping are essential tasks in such scenarios, for which 
 exploiting mobile robotic networks is extremely beneficial over using human rescuers in terms of safety and coverage. 
%
%
Teams of small autonomous agents (e.g. biologically inspired milli-robots \cite{Haldane13} or biobots), in particular,  have certain advantages over traditional platforms  since larger robotic systems may not be able to reach certain locations under the collapsed buildings. Furthermore, biobotic platforms can be particularly versatile at locomotion and navigation in unstructured and dynamic scenes due to their natural ability to crawl through small spaces.

Closely related to our work are the techniques in \textit{simultaneous localization and mapping} (SLAM) literature \cite{thrun2002robotic,slam06}. However, the majority of recent approaches assume that the exploring agents are equipped with high-bandwidth sensing devices such as cameras or depth sensors
\cite{henry2010rgb}.
On the other hand, 
mapping and localization in under-rubble environments using such platforms becomes extremely challenging due to hardware limitations and the unstructured nature of the environment. Power and computational resource constraints prohibit us from using traditional on-board imaging techniques for their localization (e.g., visual SLAM \cite{huang2011visual,henry2010rgb}). Furthermore, traditional signal propagation based localization (e.g., GPS, or computing signal strength or time of flight \cite{Ferris2007, kurth2003experimental}) may be unreliable for indoor or underground locations in cluttered environments, and odometry and inertial-based approaches (e.g., \cite{choset2005principles}) may be unreliable due to irregular conditions of the terrain.
Hence, in this work, we focus on minimal sensing strategies for robotic networks in which such sensing modalities are unavailable.


%
Since obtaining an accurate  map of the environment may not be possible in such scenarios, we propose an approach to  
construct a map that captures robust \textit{topological} features and approximates \textit{geometric} information of the environment using only
coordinate-free local interactions among the agents in a biobotic network. 
Such maps can be used  for identifying local and global homology (i.e., connected regions and obstacles) of the space, identifying homotopic trajectories connecting regions in the space, and providing an approximate and intuitive visualization of the space. Additionally, the topological information can be used as a constraint to ensure that a metric reconstruction properly captures connectivity in the space as it is done via loop-closure in traditional SLAM approaches. Moreover, instead of providing continuous control feedback to the agents, we exploit natural stochastic motion models of the biobots and their encounter information. These strategies for motion and sensing accommodate for the hardware limitations of the biobotic platform under consideration.

Methods from computational topology, on the other hand, can provide tools to    
extract topological features from point cloud data  without requiring coordinate information, which makes them more suitable for scenarios in which weak or no localization is provided.
In particular, \textit{topological data analysis} (TDA), introduced in \cite{Edel2000},  
employs tools from \textit{persistent homology} theory 
\cite{Edels08}
to obtain a qualitative description of the topological attributes 
of  data sets sampled  from  high dimensional point clouds. 

Topological frameworks have been employed in applications on sensor and robotic networks, e.g. for coverage  and hole detection in such networks,  under noisy and limited sensing with neither metric nor localization information available 
\cite{Tahbaz-Salehi2010, Ghrist, DeSilva2006,munch2012failure, kumar2015icra}.
However, these studies focus on stationary networks,  are mainly concerned about the topology of the  network rather than the  characterization of the environment itself, and bypass the geometric estimation requirement by considering a purely topological algorithm.
%
In \cite{Walker2008} persistent homology was used in order to 
compute topological invariants and 
infer global topological information of spaces from 
interacting nodes in mobile Ad-Hoc networks; yet the nodes are assumed to follow a simple mobility model restricted to a graph.  
Topological localization and mapping has been also considered lately in \cite{Mike12,Rob13,Rob12}. However, the approach in \cite{Mike12} is mostly focused on the localization problem, and 
 requires the existence of transmitters in the environment providing signals of opportunity.
The methodology of \cite{Rob13,Rob12}, 
  assume the existence of an IMU or a motion model for prediction of the agents' states, as well as fixed landmarks on the boundaries of structures (e.g. buildings) used for observation. In this paper, however,  the mapping is intended to be performed using local \textit{encounters} among the agents and not relying on external resources (signals of opportunity or known fixed landmarks), as they may be unavailable as a consequence of the emergency response scenarios under consideration.
This work is a formal mathematical framework for the approach introduced in \cite{Dirafzoon2013} and extended in \cite{Dirafzoon2014} for  mapping of unknown environments 
under limited sensing constraints, where a notion of \textit{encounter metric} is introduced based on local interactions among stochastic agents, and  tools from TDA  were exploited to extract topologically persistent spatial information such as connected components and obstacles of the environment. 
We implemented the approach of \cite{Dirafzoon2013} on a multi-robot network,  the \textit{WolfBots} \cite{Dirafzoon2014}.
In this paper, we expand the metric estimation procedure of \cite{Dirafzoon2013} in order to construct more accurate geometric point clouds of the environment, and analyze the precision of proposed methodology from a \textit{manifold learning} \cite{mani-vis} perspective. We derive bounds on the metric estimation error, and present conditions for which asymptotic convergence to geodesic distances holds. 
Furthermore, we present a topological framework to infer persistent features in the map of the environment, and analyze its stability under geometric uncertainty. 
This work, can be used as building blocks for mapping of larger scenarios by herding the network using an aerial vehicle sweeping the environment. 

The remainder of the paper is organized as follows: 
A concise background on the mathematical tools used throughout the paper is provided in section \ref{sec:Background}.
Section \ref{sec:ProblemStatement} describes the problem under study including models and assumptions for the agents in a biobotic network, followed by 
an overview of our  approach in section  \ref{sec:Method}. Geometric estimation methodology including asymptotic convergence analysis is studied 
in  section \ref{sec:Metric}, and topological estimation and its stability analysis 
are discussed in section  \ref{sec:topInf}. 
Section \ref{sec:SimulationResults} illustrates and validates the approach via numerical simulation experiments, and finally, conclusion and future work are provided in section \ref{sec:conclusion}.

\section{Mathematical Background} 
\label{sec:Background}
\begin{figure*}[tb]
	\centering
	\includegraphics[width=0.95\linewidth]{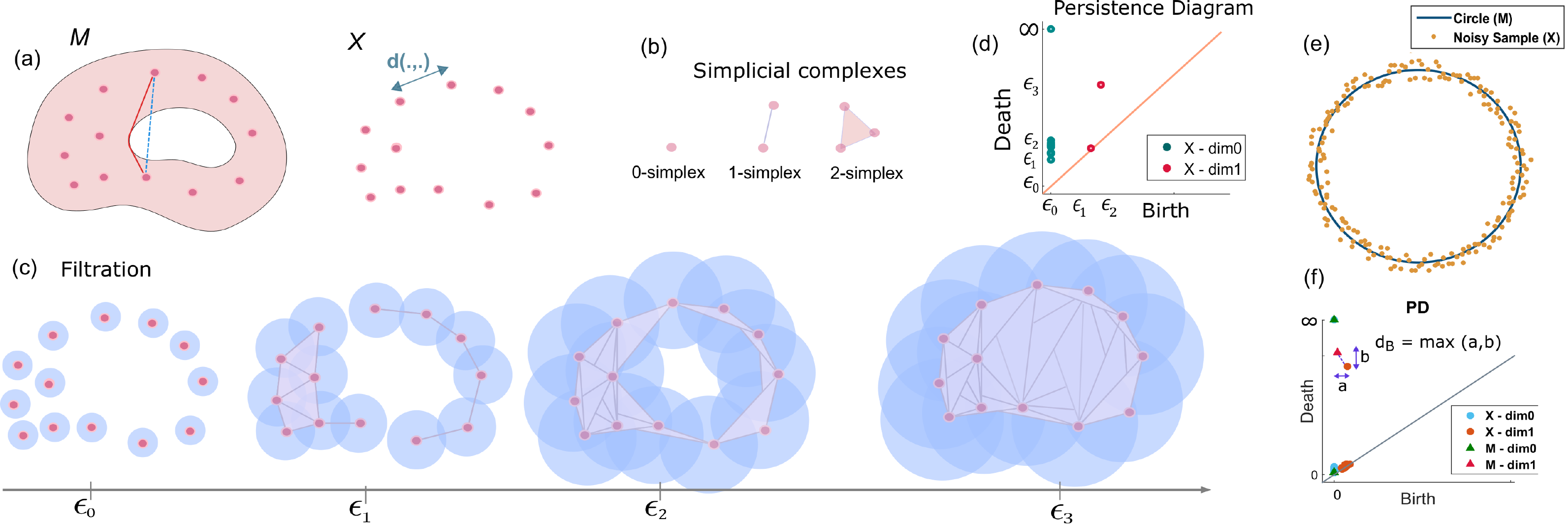}
	\caption{ {Topological persistence: (a) An example of a 
			space $ \M $ with a sampled point cloud $ X $, (b) illustration of simplicial complexes, (c) a filtration of simplicial complexes over $ X$, (d) the corresponding persistence diagram with $\pd_0(X) $ in green, and  $ \pd_1(X) $ in red, highlighting persistence of connected components and holes over the scale. Plots (e) and (f) illustrate the stability of persistence: (e) a circular space $\M$ with noisy samples $X$, (f) persistence diagram of $\M$ and $X$ and their bottleneck distance.}}
	\label{fig:filtr}
\end{figure*}
\subsection{Manifold learning}
%
%

Consider a manifold $\M$ whose structure and features are of  interest (e.g. an unknown environment to be mapped). 
$\M$ usually is not directly accessible but  a  finite sampled version of it, ${\X = \{x_1, \ldots, x_n\},}$ can be used for computations. $\X$ can be represented as a \textit{point cloud},  a  set of  points equipped with a \textit{metric}, defined by pairwise distances between the points.
The metric 
can be obtained directly using the coordinates of the points, or by estimation of pairwise distances independently from their (possibly unknown) coordinates (see Fig. \ref{fig:filtr} (a)).

Given a finite set of  sampled data points $\X$, \textit{manifold learning} is referred to the set of algorithms that try to discover the intrinsic low dimensional structure of the embedding manifold $\M$ of the data set. 
A review of different manifold learning approaches can be found in \cite{lin2008riemannian}.  
\textit{Isomap}\cite{Tenenbaum2000}, in particular, 
learns the underlying global geometry of the manifold captured given only  sampled data points  by preserving the \textit{geodesic distances}. 
The geodesic distance between two points is the length of the shortest curve $\gamma$ (geodesic) that connects the two points on $\M$, i.e. $
\dm (x_i, x_j) =\inf_\gamma \{\text{length}(\gamma)\}. 
$  As an example, on a sphere, the geodesic  distances are obtained from the great circles.
The main idea behind Isomap is to approximate geodesic distances on the manifold by the shortest path distance on the neighborhood graph of the samples taken from the manifold, using measured/estimated local metric information.  
Isomap is guaranteed to asymptotically recover geometric structure of a manifold given a dense enough sampled data set\cite{Tenenbaum2000}. 
\subsection{Topological Data Analysis (TDA)}  
TDA, on the other hand, provides a framework for description of topological features in a point cloud in a coordinate free manner, while being robust to metric perturbations and noise. It employs persistent homology theory
 to extract the prominence of such features
in terms of a compact multi-scale representation
 called \textit{persistence diagrams} \cite{Edels10}.
\subsubsection{Simplicial Complexes}
A standard method to analyze the topological structure of a point cloud  is to map it  into  combinatorial objects called {\it  simplicial complexes}. 
Given a set of samples $\mathcal{X}$, a $k$-simplex is defined as a set 
$
\{x_1, x_2, \ldots, x_{k+1}\},  \; x_i \in\mathcal{X},  \; \forall i \; \text{and} \;  i \neq j,  \forall i,j.  
$
For example, a $0$-simplex is a vertex, a $1$-simplex is an edge, and a $2$-simplex is a triangle. 
A finite collection of simplices is called a \textit{simplicial complex} if for each simplex each of its faces are also included in the collection. 

Fig.~\ref{fig:filtr}(b) shows an example of 0,1,2-simplices constructed on points drawn from $\M$. 
One way to build these complexes is to select a scale $\epsilon$, place balls of radius $\epsilon$ on each vertex, and construct simplices based on their pairwise distance relative to $\epsilon$. A computationally efficient complex, called the \textit{Vietoris-Rips complex} $\mathcal{R}(\X, \epsilon)$, 
consists of simplices for which the distances between each pair of its vertices are at most $\epsilon$. In other words, given pairwise distances $d$ over $\X$,  any subset $ \X' \subset \X $ is a simplex in $\mathcal{R}(\X, \epsilon)$ if and only if
$
d(x_i, x_j) \leq \epsilon, \; \forall x_i, x_j \in \X'. 
$
Note that the construction of Vietoris-Rips complex only requires the knowledge of pairwise distances among the sampled points. 
\subsubsection{Persistent Homology}
Topological invariants of a  space $\M$ are mappings which aim to classify equivalent topological objects into the same classes.  Homology (a type of topological invariance) can be summarized as a compact representation in the form of so-called  Betti numbers, which are the ranks of the \textit{homology groups}. The $n$-th Betti number, $ \beta_n $ is a measure of the number of $n$-dimensional cycles in the space (e.g. $\beta_0$ is the number of connected components and $\beta_1$ is the number of  \textit{holes} in the space).
For a dense enough sample, one can find an interval for the values of $\epsilon$, for which constructed simplicial complexes belong to the same class of topological invariants as the space $\M$. 
However, finding the proper values for $\epsilon$  is a strenuous job. 

On the other hand, \textit{persistent homology} theory considers 
construction of a sequence of such complexes for the  sampled point cloud over multiple scales $\epsilon$, called a \textit{filtration},  and extracts  geometrical and topological features of the dataset at specific scales. 
A filtration ${X}(\epsilon)$ can be obtained by increasing $\epsilon$ over a range of interest,  with the property that 
if  $ t< s, $ then  ${X}(t)\subset {X}(s)$. 
%
%
Persistent homology computes the  values of  $\epsilon$ for which the classes of topological features  appear ($b^i_n$) and disappear ($d^i_n$) during filtration,   referred to as  the \textit{birth} and \textit{death} values of the $i$-th class of features in dimension $ n $. 
This information is encoded into persistence intervals $[b^i_n, d^i_n]$, or  a compact representation 
called \textit{persistence diagram} \cite{Edel2000}. 
The persistent diagram $\pd_n ( {\X} )$, 
is a  multi-set of points $ (b^i_n, d^i_n) $  
that summarizes how topological features of the point cloud vary over the scale $\epsilon$.
Algorithms for computation of persistent homology  can be found in 
\cite{Edel2000}.
An example of a topological space $\M $ together with sampled data ${X} $ and corresponding filtration over $ \epsilon $ is shown in Fig.~\ref{fig:filtr} (a-c). 
In the persistence diagram in Fig.~\ref{fig:filtr}(d), features corresponding to dimension 0  and dimension 1 
are shown in  green and red colors, respectively. The space $ \M $ can be described topologically as having 1 connected component and 1 hole, meaning that $ \beta_0 = 1$ and $\beta_1 = 1 $ for this example. Although there exist several topological features in the $ \pd $, only 2 of them can be distinguished from the rest based on their distant from the diagonal. These points correspond to the \textit{persistent features} with longer persistence intervals, referred to as \textit{topological signals}, while the rest are short-lived features that are referred to  as \textit{topological noise} \cite{Horak2009}. 
Hence one can infer the existence of one connected component and one persistent hole from the $ \pd $.
\subsubsection{Stability of Persistence Diagrams}
One of the  fundamental properties of persistence diagram is its \textit{stability}\cite{Cohen-Steiner2007}, which means a sufficiently small change in the metric results in a small change in the persistence diagram. 
There are variations of stability results based on different metric definitions. 
For a metric space $(\M, d)$ (a set $\M$ together with a metric $d$), 
the \textit{Hausdorff distance} between these two of its subsets $\X$ and $\Y$,  $d_H(\X, \Y)$,  is defined as the maximum distance from a point in one set to the closest point in the other set:
\beq
d_H(\X, \Y) =  \max \{  \max_{x \in \X} \min_{y \in \Y} d(x, y) ,\, \max_{y \in \Y} \min_{x \in \X} d(x,y)\}
\eeq
Moreover, for two metric spaces $(\X, d_1)$ and $(\Y, d_2)$ their 
\textit{Gromov-Hausdorff} distance $d_{GH}((\X ,d_1), (\Y ,d_2))$ is defined as the infimum of their Hausdorff distance over all possible isometric embeddings of these two spaces into a common metric space  \cite{chazal2009gh}.
In order to compare persistence diagrams, a distance metric called the \textit{Bottleneck distance}, is defined for two diagrams $\pd_1$ and $\pd_2$ as\cite{Cohen-Steiner2007}: 
\begin{equation}
\db 
(\pd_1, \pd_2) = 
\inf_{\phi} \, \sup_{a \in \pd_1} \|a - \phi(a) \|_\infty,
\end{equation}
where the infimum is taken over all bijections $ \phi: \pd_1 \rightarrow \pd_2 $ and
$\|.\|_\infty$ denotes supremum–norm over sets.
%

For two compact metric spaces  $(\X, d_1)$ and $(\Y, d_2)$, the persistence diagram is \textit{stable} in the sense that \cite{chazal2009gh}:
\begin{equation}\label{eq:st-gh}
\db(\pd_{\X}, \pd_{\Y}) \leq 2 . \, d_{GH} \left( (\X , d_{1}),   (\Y, d_{2})\right)
\end{equation}
Furthermore, if $\X$ and $\Y$ are embedded in the same space $(\M, d_\M)$ then (\ref{eq:st-gh}) holds for the Hausdorff distance $d_H$ instead of $2d_{GH}$. 
%
Fig.~\ref{fig:filtr} (e) shows a circular space $\M$ along with noisy samples $X$ drawn from it. The corresponding persistence diagrams $\pd_\M, \pd_\X$ are presented in  Fig.~\ref{fig:filtr}(f), which are not much different. Here the bottleneck distance can be described as the $L_\infty$ norm between the points corresponding to the holes in each space as shown in the plot.

\section{Problem Formulation} \label{sec:ProblemStatement}

Consider a network of biobotic agents moving stochastically within a bounded domain of interest $ \M \subset \mathbb{R}^2$.
We restrict our analysis to a 2D environment for simplicity while the techniques introduced do not rely on this assumption. Each  agent is distinguished by its unique  ID belonging to the set $\I = \{1, \ldots, n\} $,  and its  motion dynamics in the bounded space mimics the movement model of a biobotic insect. The movement model of the biobotic insect can be described by two types of motions: \textit{natural motion} and \textit{controlled motion}, which will be described in the following subsection. 
\subsection{Motion Model of the Biobotic Networks }\label{subsec:model}
\textit{{Natural Motion (NM)}}: 
The {natural motion} model of the biobotic agents is adopted from the probabilistic  motion model of cockroaches in bounded spaces described in \cite{Jeanson2003}, described by their \textit{individual} and \textit{group} behaviors.
The  individual movement of  cockroaches can be characterized mainly by two behaviors, namely 
\textit{correlated random walk (CRW)} 
and \textit{wall following} (WF).  
Cockroaches, in their natural mode, are known to  move according to a CRW model when they are far enough from the boundaries of the domain, and perform WF behavior when they detect the edges of the environment using their antennas \cite{Jeanson2003}. 
During WF, they switch back again to CRW motion towards the interior of the domain after some stochastic amount of time.   Moreover, it is known that during their CRW or WF motion, the insects  probabilistically \textit{stop} for some period of time and then continue their movement \cite{Jeanson2003}. We refer to the insects being in a \textit{static (S)} mode when they stop. 

The CRW is modeled as piecewise linear movements with fixed orientation, characterized by line segments $\li_i$, 
interrupted by isotropic changes in direction $\theta_i$, and constant average velocity of $v_m$.
The lengths of  line segments $\li$  has an exponential distribution  with characteristic length $\li^*$:
\beq\label{eq:crw}
p(\li) = {\li^*}^{-1} e ^ {-\li \,\,/ \,\,\li^*},
\eeq
Changes in orientation also triggered by collision detection with other agents or obstacles in the scene. 
Fig. \ref{fig:front}(c) shows an example of CRW motion with few line segments $ \mathpzc{l}_i $, with angular re-orientations of $\theta_i$. 
Cockroaches also manifest a group behavior, \textit{aggregation}, when they interact with each other \cite{Jeanson2005}. 
%
In this paper, however, we focus on the CRW behavior for simplicity. Hence our model of natural motion will be reduced to a two state graphical model \textit{switching} between CRW and S. A. Dirafzoon et al \cite{Dirafzoon2013} explored the effect of WF behavior on the exploration and mapping process.
Note that although this model is particularly developed and studied for German cockroach \textit{Blattella germanica}, many animals (in particular other species of cockroaches) are known to display similar motion behaviors with possibly different parameters for CRW or \textit{Levy Walk} \cite{Plank2013}.

 \textit{{Controlled Motion (CM)}}:
  The remotely controlled biobots
 are equipped with 
 system-on-chip based 
 commands of the following types: \textit{turn left}, \textit{turn right}, \textit{start} or \textit{stop} motion. 
 In this paper, we only consider start and stop  commands  in order to 
 switch the mode of the agents from CRW to S and vice versa. 

\subsection{Sensing and Processing }
The  sensing and communication model for the agents is inspired by limited sensing capabilities of the biobotic insects, consisting of a wireless transmitter and receiver provided by the system-on-chip based ZigBee enabled backpacks attached to their bodies \cite{latif2012}. Each agent is assumed to have a limited proximity sensing capability, and can identify and communicate with other agents  within a detection radius $r_d$, as shown in Fig.~\ref{fig:front}(d). The agents are able to record their \textit{encounters} with each other and the corresponding times as \textit{encounter events}. 
Furthermore,  the nodes are  able to report their status as being in a CRW or S state. 
Local encounters are the only piece of information provided, and no coordinates or other localization information is available. Note that odometry and inertial measurements are considered to be too unreliable due to the uneven and unstructured terrains present in our application scenario.  
We do not consider any external obstacle detection/collision mechanism as we rely on  the insects' ``built-in'' instinct for collision detection and their natural change of direction in case of collision.
\begin{figure}[tp!]
	\centering
	\includegraphics[width=1\linewidth]{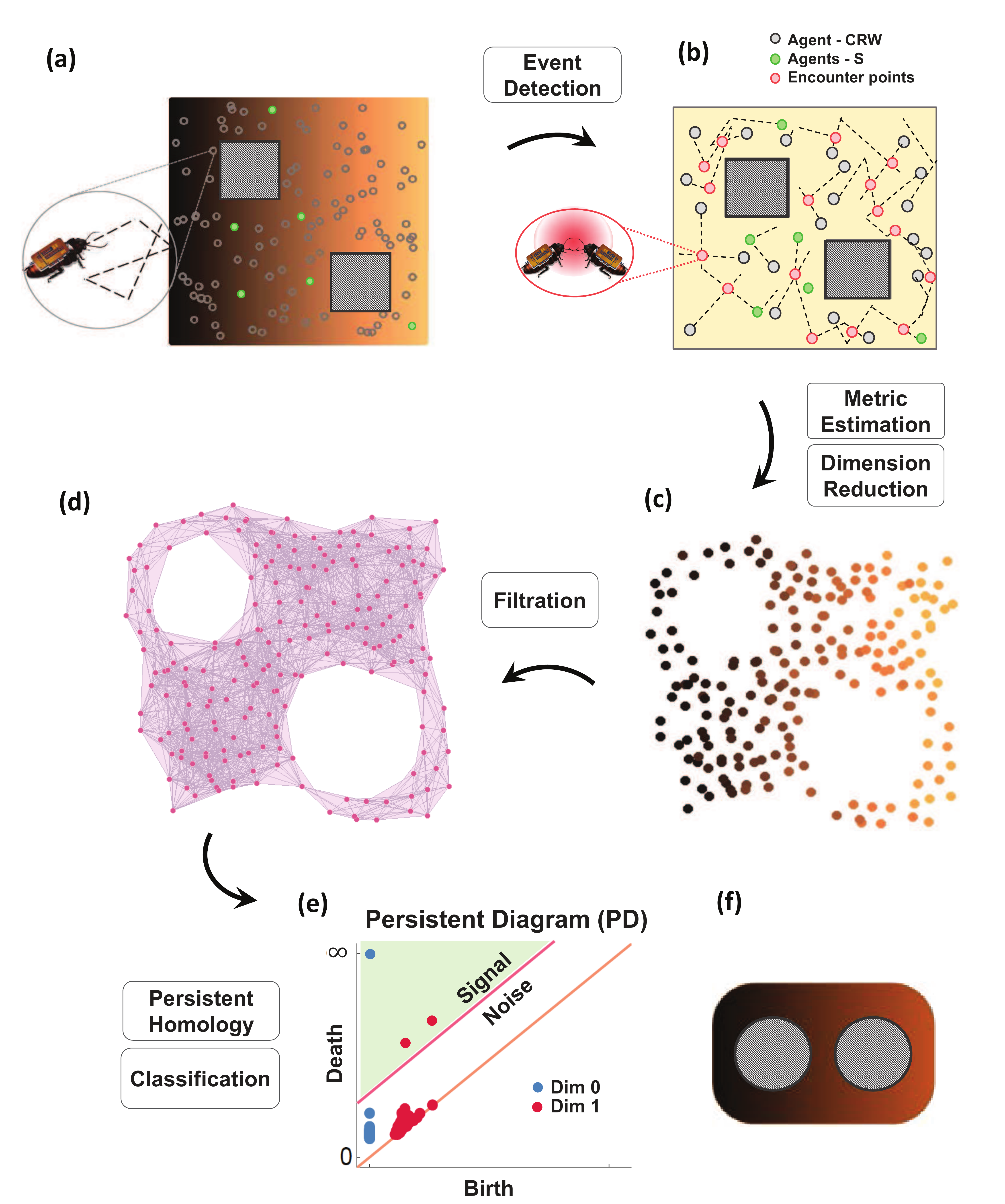}
	\caption{\small{Overview of our methodology: 
			(a) an  environment with two obstacles and a snapshot of the location of 100 agents (gray circles), (b) a subsample of collected coordinate-free encounter nodes (pink circles) over which a metric is estimated, (c) the point cloud obtained from projection of the encounter metric (d) a simplicial Rips-complex representation, (e) persistence diagram of the complex highlighting connected components in blue and holes in red dots classified as signals and noisy features, and (f) a topological sketch of the environment.
		}}
		\label{fig:top-map}
	\end{figure}
\subsection{Objective}
Given this minimal sensing scenario, our goal is to learn the underlying geometry and infer the topology of the unknown environment as the swarm of agents efficiently explore the domain. 
As the swarm of agents 
share information about their local
\textit{encounters} during exploration, a coordinate-free  map of the scene is built, that consists of geometric point clouds representing estimates of distances between the sampled points, and a topological summary of the environment (e.g. connected components and obstacles). We refer to such a map as \textit{Geometric-Topological Persistent Map} (GTP-Map) of the environment. We measure the quality of geometric estimation via computation of errors in geodesic distance estimated and quality of topological inference via errors in the topological features in the persistence diagrams and the inferred ranks of the homology of the space. 

\section{Methodology}\label{sec:Method}
A high level  overview of our approach is illustrated in Fig.~\ref{fig:top-map}. 
A physical environment with 2 holes (corresponding to obstacles) is shown in Fig.~\ref{fig:top-map}(a) with agents whose motion mimics the biobots movement model. 
The objective is to estimate a 2D embedding which best represents intrinsic geometry and topological features of the environment using local interactions among the agents. 
This task is accomplished by the following stages: (1) exploration and event (data) collection, (2) metric reconstruction, and (3) topological feature extraction. 

%
%
\textit{{Exploration and Event Collection}}: 
The exploration  is initiated with  a  \textit{dispersion} stage, where  all nodes start their motion
 in a CRW status until dispersed throughout the environment,
while maintaining their connectivity.
This is followed by a \textit{landmark selection} stage, where a small percentage of the nodes are selected as static  \textit{landmarks}  using a modified  \textit{maxmin} landmark selection approach \cite{Silva2004}, and are commanded to switch to S mode. Given the adjacency graph of the communication, we use its hop distance to select landmarks iteratively by maximizing the distance from the previous set of landmarks. 
If the graph is not connected, the  landmark agents are selected randomly from each cluster with probability equal to the percentage of agents in each cluster. 

%
The encounters among the agents (Fig.~\ref{fig:top-map}(b)) are recorded as data points, 
over which a distance metric is estimated after it is pre-processed. 
The estimated point cloud
is projected to 2D space using a dimension reduction technique (Fig.~\ref{fig:top-map}(c)). 
Persistent homology extracts the persistence diagram for the simplicial filtrations of the point cloud (Fig.~\ref{fig:top-map}(d)) which will be used for topological inference. Details on metric  and topological estimation are provided in the following sections.

\section{Geometric Estimation and Analysis}
\label{sec:Metric}

 \subsection{Metric Estimation}

An \textit{encounter event} $\e_i$ takes place between two agents if their corresponding 
positions fall within a distance of $r_d$, 
and is recorded by an agent as a tuple:
\begin{equation}
	\e_i = (t^1_i, t^2_i, \id_i^1, \id_i^2),
\end{equation}
where 
$[t^1_i,t^2_i]$
 denotes the time interval of the encounter, and $ \id_i^1 $ and $ \id_i^2$ represent the IDs of the two agents encountering. 
Define   $t_i  =  (t^1_i+ t^2_i)/2$ as the  center of the time interval, and   $\id(\e_i ) = \{\id_i^1, \id_i^2\}$  as the ID set of the event $\e_i$. 
 %
Assigned to each event $\e_i$ is a coordinate 
 $p_i \in \M$, that 
represents the center of mass of the set of  
location at which the agents with $\id_i^1$ and $\id_i^2$ met. 
We denote the set of all events as  $\E = \{\e_i\}$, and the set of corresponding coordinates as $\Ps  = \{p_i\} \subset \M$. 

Furthermore, define the \textit{landmark set} $\La  = \{l_k\} \subset \I$ as the set of IDs of the agents that are assigned to be static (in S mode). 
Given the set  $\La$, then the set $\C_k \subset \E$ of all \textit{landmark events} corresponding to the $k$-th landmark is defined as 
\beq
\C_k = \{\e \,\, | \,\, l_k \in \id(\e)\}. 
\eeq
Note that $\C_k$ represents a cluster of events, which is referred to as a \textit{community} in a graph; so we refer to the set $\C_k$ as the $k$-th \textit{landmark community}. 
For each landmark community $\C_k$, define its coordinate $q_k$ as the center of mass of the positions of all the events in the cluster. We denote the set of all landmark communities as $\Cs= \{\C_k\}$, and the corresponding coordinates as $\Ls  = \{q_k\} \subset \M$.
 
%
The sets $\E$ and $\Cs$ can be considered as unlocalized samples gathered from the environment, which can be projected into sampled subspace $\Ps$ and $\Ls$
drawn from $\M$. 
 We construct distance metrics on $\E$ and $\Cs$ that approximate pairwise distances close to the geodesic distances in $\Ps$ and $\Ls$ induced by the manifold distances $d_\M: \M \times \M \to \mathbb{R}^+$. The idea is to approximate geodesic distances 
  by the shortest path distance on the neighborhood graph of the samples taken from the manifold.

Construct  an undirected weighted graph $\G$ with vertices corresponding to the events $\e_i$, denoted as the \textit{encounter graph}. Given
$\G$,  
for any two vertices $\e_i$ and $\e_j$, include an edge $\e_{ij}$ if  they have a common agent
involved in both events, i.e. $\id(\e_i) \cap \id(\e_j) \neq \emptyset$. The corresponding weight 
$w_{ij}$ will represent pairwise distances to be calculated as follows. 
For $\e_i$ and $\e_j$, define the pairwise \textit{ encounter time metric} $ \de $ as
%
$
\de(\e_i, \e_j) =
	| t_i - t_j |
\label{eq:time-dist}
$
if $\e_i$ and $\e_j$ are connected. 
%
%
%
Note that, for connected nodes $\e_i$ and $\e_j$, the Euclidean distance between the corresponding coordinates $p_i$ and $p_j$ is upper bounded as 
$ 
\| p_i - p_j \|_2 \leq  v_m\,.\,| t_i -  t_j |. 
$
Provided that the agents were moving only on straight lines and there were no obstacles in the field, this upper bound could have  precisely  approximated the Euclidean distance. However, due to CRW motion of the agents and existence of the obstacles, the accumulated errors of the pairwise distances over time will cause large drifts in metric estimation.  
%

To reduce the amount of such drifts, we furthermore improve our graphical model by incorporating the fact that landmark nodes are static. Consider again the set  $ \La $ of the landmark nodes. 
 Then, for two events $\e_i$ and $\e_j$, if there exists a landmark community such that 
    $\e_i, \e_j \in \C_k,$
    they had to occur geometrically at nearby locations. 
Therefore, the corresponding weight $w_{ij}$ is set to $0$ in this case, and we define a pairwise \textit{encounter metric} $ \del  $ as 
\beq
 	\del (\e_i,\e_j) =
 	\begin{cases}
 		0, &  \e_i, \e_j \in  \C_k \text { for some } k,\\
 \de(\e_i, \e_j) , & \text{otherwise}.\
 	\end{cases}  \label{eq:pdist}
\eeq

To proceed with the construction of geometric point cloud, we perform a graph shortest path algorithm on the graph $ \G$, and define our \textit{encounter metric} on the nodes in $\G$ as: 
\begin{equation} 
\dgel (\e_i,\e_j) = \min_{\p} \{ \del (\x_0, \x_1) + \ldots  + \del (\x_{p-1}, \x_p)  \}, 
	\label{eq:gdist}
\end{equation}
where $\p =(\x_0=\e_i, \x_1, \ldots, \x_p = \e_j)$ denotes a path along the edges of graph $\G$ connecting $\e_i$ to $\e_j$. The shortest paths and their corresponding lengths can be found from Dijkstra's or Floyd's shortest path algorithms \cite{csrl01}. The graph distances  $\dge$ are defined similarly.

Graph $ \G $ consisting of nodes corresponding to the events $\e_i$ and  edge weights set as $w_{ij} = \dgel (\e_i,\e_j)$, 
 is referred to as the \textit{encounter metric graph}, which approximates the geodesic distances over $\M$ from the sample points.  
We  refer to the set of nodes in $\G$ together with the metric $\dgel$ as the  \textit{encounter point cloud} of $\M$. 
The estimated point cloud lies in a higher dimensional space although its intrinsic geometry can be embedded in a lower dimensional spaces. We make use of classical \textit{Multi-dimensional Scaling} (MDS) \cite{mds} for dimension reduction in order  to find a 2D embedding of the point cloud data. MDS, is typically used for visualization of dissimilarity matrices by calculating a set of of coordinate whose configuration minimizes a loss function \cite{mds}. 


\subsection{Metric Analysis} 
\label{sec:metric-analysis}

In the following, we analyze and discuss convergence and asymptotic behavior 
of the proposed metric for approximation of geodesic distances, and investigate assumptions and conditions on sampling and graph connectivity under which $ \dgel $ is guaranteed to recover the true geometry of $\M$. 

For events $ \e_i$ and $\e_j $, define $\dm$ as the distances between their corresponding coordinates, $p_i$ and $p_j$, i.e. $ d_\M(\e_i, \e_j) = d_\M(p_i, p_j) $. 
 Moreover, for a point $p \in \M $ and an event $\e_i$, we define the pairwise distance $\dm(p, \e_i)  = \dm(p,p_i)$. 
For an event $ \e_i $ and a landmark  community $ \C_k $, define their pairwise distance as $ \dgel(\e_i, \C_j) = \min_{\e \in \C_j} \dgel(\e_i, \e)$. Accordingly, 
for two landmark communities $ \C_i, \C_j, $  set the \textit{encounter metric for communities} $\dgel(\C_i, \C_j)$ as 
the shortest path distance from one community to another.

In the following we will show that if there exist enough number of encounters among the agents and the encounter graph is sufficiently dense, the encounter metric for communities will converge to the manifold distances between their corresponding positions in $\M$. Furthermore, the encounter metric for all pair of nodes in $\G$ will converge to the manifold distance as the number of landmarks increases.

In our analysis, we make an assumption of unit linear velocity for CRW motion, i.e. $ v_m = 1,$ for the sake of simplicity. The results can be easily extended for  a general case with a scaling factor of $1/ v_m $. Moreover, we neglect the effect of the detection radius $r_d$ on the obtained bounds; yet it can be shown that it will introduce a bias term proportional to $r_d$ in the errors. Under these assumptions, we can identify an event $\e_i$ with a single point $p_i$ and time $t_i$. For simplicity, we abuse our notation and redefine $\e_i = (p_i, t_i) \in \M_T :=\M \times T$, where $T$ represent the time domain, and define the projections $\Pi_\M:\M_T \to \M$ and $\Pi_T:\M_T \to T$. We further define a metric on the product space $\M_T$ as
\beq \label{eqn:def_dmxt}
  \dmxt(\e,\e') = \inf_{\Gamma \in A(\e,\e')} S_\Gamma,
\eeq
where $A(\e,\e')$ is the set of continuous functions of the form $\Gamma(s) = (\gamma_\M(s),\gamma_T(s))$ such that $\Gamma(0)= \e$ and $\Gamma(S_\Gamma) = \e'$, $|\nabla_\M \, \gamma_\M (s) | = 1$ and $|\frac{\partial \gamma_T}{\partial t}(s) | = 1$. This set of trajectories correspond to curves in $\M$ traveled at constant velocity that have been lifted to $\M_T$.

%
%

In order to analyze the asymptotic behavior of our approach, we assume that the set of trajectories and events become dense enough such that we can approximate any of the geodesics trajectories in $\MT$.
Let $\delta_e$ and $\varepsilon$ be positive real numbers. We require the following conditions to hold:

\begin{assumption}[Sampling Conditions]\label{asum:1}
	\begin{enumerate}[(i)]
		\item For every point $  \e \in \M_T$, there exists a sampled event $\e_i$ such that $\dmxt (\e, \e_i) \leq \delta_e $.
		\item For any two neighboring events $\e_i$ and $\e_{j}$ such that $\dmxt(\e_i, \e_{j}) \leq \varepsilon$, there exists a $\tau$ such that $ \de (\e_i , \e_{j}) \leq \tau$, where $\tau \to \epsilon$ as the number of agents increases to infinity.
	\end{enumerate}
\end{assumption}
\begin{assumption}[Graph Connectivity Condition]\label{asum:2}
	The graph $\G$ contains all edges for which $ d_{\M_T}(\e_i,\e_{j}) \leq \varepsilon$. 
\end{assumption}

Condition (i)  corresponds to the \textit{$\delta_e$-dense encounter sampling} condition for the events, which is guaranteed as the number of agents or the running time increases. Condition (ii) can be guaranteed by the motion models considered. As the number of agents  increases, each path connecting nearby events will look more like a straight path (due to the CRW model under consideration).

\begin{lemma}\label{lem:1}
	For any two encounter events $\e_i$ and $\e_j$ we have:
	\begin{eqnarray}
	\dmxt(\e_i, \e_j) \leq \dge (\e_i, \e_j). \label{eq:dm-dge}
	\end{eqnarray}
	
	\begin{IEEEproof}
		First, we note that given neighboring events $\e_i$ and $\e_j$ in $\G$ then $\dmxt(\e_i,\e_j) \leq \de(\e_i,\e_j)$ since $\dmxt$ is defined as the infimum over all trajectories of constant velocity, and the trajectory between $\e_i$ and $\e_j$ is one particular realization. Furthermore, the trajectory defined by the path $\p =(\x_0=\e_i, \x_1, \ldots, \x_p = \e_j)$ associated with the shortest path in $(\G,\de)$  corresponds to one particular realization of the set of trajectories considered in the definition of $\dmxt$; hence, we have the desired inequality.
	\end{IEEEproof}
\end{lemma}

\begin{lemma}\label{lem:2}
	For two encounters $\e_i $ and $\e_j$,  under sampling Assumption 1 and graph connectivity Assumption 2, 
	if $ 4 \delta_e < \varepsilon$ then:
	\begin{equation} \label{eq:dgm}
	\dge (\e_i, \e_j) \leq (1+\lambda_1) \cdot \lambda_2 \cdot \dmxt(\e_i, \e_j), 
	\end{equation}
	where  $ \lambda_1 = 4 \delta_e / \varepsilon$ and $\lambda_2 = \tau/\varepsilon$.
\end{lemma}		
\begin{IEEEproof}
	Let $\Gamma(s):=\left(\gamma_\M(s),\gamma_T(s)\right)$ for $s \in [0,S_\Gamma]$ be an arbitrary curve in $A(\e_i,\e_j)$ as defined in Equation (\ref{eqn:def_dmxt}). We decompose this curve into arcs with lengths $s_k$ by defining a sequence of points $\Gamma_0 = \e_i, \Gamma_1, \ldots, \Gamma_n,\Gamma_{n+1} = \e_j$ along $\Gamma$ such that $S_\Gamma = \sum_{k=0}^n{s_k}$ with $ s_k  = \varepsilon - 2\delta_e $ for $ k = 1, \ldots,  n-1 $, and $(\varepsilon - 2\delta)/2  \leq s_0, s_n \leq \varepsilon - 2\delta_e$.

	Based on condition (i) of Assumption 1, for each point $\Gamma_k$ there exist an encounter event $\x_k$ within distance $\delta_e$ from $\Gamma_k$. Additionally, for each pair $ (\x_k, \x_{k+1}) $ in the path $\p = (\e_i, \x_1, \dots , \x_{n}, \e_j)$ we have 
	$
	\dmxt(\x_k, \x_{k+1}) \leq \dmxt(\x_k, \Gamma_k) + \dmxt(\Gamma_k, \Gamma_{k+1}) + \dmxt(\Gamma_{k+1}, \x_{k+1}) \leq s_k + 2 \delta_e = \varepsilon
	$ for $k = 1,\ldots,n-1$, which implies that there is an edge between $\x_k$ and $x_{k+1}$ in $\G$ by Assumption 2. Then, by condition (ii) in Assumption 1, $\de(\x_k,\x_{k+1}) \leq \tau = \lambda_2 \cdot \varepsilon = \lambda_2 \cdot \varepsilon \cdot s_k / (\varepsilon - 2 \delta_e)$, where the last equality follows from the definition of $s_k$. Similarly, one can show that $\de(\e_i,\x_{1}) \leq \lambda_2 \cdot \varepsilon \cdot s_0 / (\epsilon - 2 \delta_e)$ and $\de(\x_{n},\e_j) \leq \lambda_2 \cdot \varepsilon \cdot s_n / (\epsilon - 2 \delta_e)$.
	
	By summing all distances along the path $P$, we have that $\dge (\e_i, \e_j) \leq \de (\e_i, \x_1) + \de(\x_1, \x_2) +  \dots + \de(\x_n, \e_j) \leq  \lambda_2 \cdot \varepsilon \cdot  S_\Gamma / (\varepsilon - 2\delta_e)$. Using the fact that $ 1 / (1 - x) < 1 + 2x $ for positive $ x < 1/2 $, one concludes $\dge (\e_i, \e_j) \leq \lambda_2 \cdot S_\Gamma \cdot (1 + 4 \delta_e / \varepsilon) $ for $4\delta_e < \varepsilon$. 
	Taking the shortest path among all paths between $\e_i$ and $\e_j$ we have the desired result. 
\end{IEEEproof}

Combining the above lower and upper bounds on $\dgel (\e_1, \e_2)$, we can conclude the main result for the convergence of distances as follows.

\begin{theorem}\label{thm:1}
	For any  pair of events  $ \e_i, \e_j$, provided that Assumptions 1 and 2 hold, and  $ 4\delta_e < \varepsilon $ then
	\begin{equation}
	\dmxt(\e_i, \e_j)   \leq   \dge(\e_i, \e_j)   \leq (1+\lambda_3) \cdot \dmxt(\e_i, \e_j),   
	\end{equation}\label{eq:thm}
	where $\lambda_3 = \tau/\varepsilon - 1 + 4\delta_e \tau / \varepsilon^2$.
\end{theorem}

Landmark communities are the set of events associated with a particular landmark at a fixed location, so we can represent them in $\MT$ as the set $\mathcal{S}_k = \Pi_\M^{-1} (p_k)$ where 
$p_k  \in \M $ 
is the physical location of landmark $l_k$. Hence, we can define a new metric space $\MTL$ which is $\MT$ with the proper identification along the set of landmark communities $\{\C_k\}$. We can show that this new space has a similar metric to the original space.

\begin{lemma} \label{lem:dmxtldm}
	Given two landmark events $\e_k \in \mathcal{S}_k$ and $\e_{k'} \in \mathcal{S}_{k'}$, where $p_k$ and $p_{k'}$ are the location of landmarks $l_k$ and $l_{k'} \in L$, then
	\beq
	\dmxtl(\e_k,\e_{k'}) = \dm(\e_k,\e_{k'}).
	\eeq
\end{lemma} 

\begin{IEEEproof}
	Note that the definitions of $\dm$ and $\dmxtl$ both require curves $\gamma_\M(s)$ parameterized by arc-length in $\M$. However, $\dmxtl$ would also impose some additional conditions on $\gamma_T(s)$ based on the structure in $\M_T^L$. Therefore, $\dmxtl(\e_k,\e_{k'}) \geq \dm(\e_k,\e_{k'})$ since the infimum for $\dm$ is over a bigger set.
	
	Furthermore, given that $\gamma^*_\M(s)$ is a geodesic in $\M$ with arc-length $S_\Gamma^*$ joining $p_k$ and $p_{k'}$, then $\Gamma(s)=(\gamma^*_\M(s),t_k+s)$ is a valid trajectory in $\M_T^L$ joining $\e_k$ and $\hat{\e}_{k'} = (p_{k'},t_k+S_\Gamma^*).$ This means that $\dmxtl(\e_k,\hat{\e}_{k'}) \leq \dm(\e_k,\hat{\e}_{k'})$. However, since the $\e_{k'}$ and $\hat{\e}_{k'} \in \mathcal{S}_{k'}$ then they are both identified as the same point in $\M_T^L$. Hence, $\dmxtl(\e_k,\e_{k'}) \leq \dm(\e_k,\e_{k'})$, which concludes the proof.
\end{IEEEproof}

\begin{lemma} \label{lem:dmxtldm2}
	Given two events $\e_i$ and $\e_j$, then
	\beq \begin{array}{l}
	\dmxtl(\e_i,\e_j) = \min\left(\dmxt(\e_i,\e_j),\right. \\
	\left. \quad \inf_{l_k,l_{k'} \in L} \{ \dm(\e_i,\mathcal{S}_k)+\dm(\mathcal{S}_k,\mathcal{S}_{k'})+\dm(\mathcal{S}_{k'},\e_j) \}\right).
	\end{array}
	\eeq
\end{lemma} 

\begin{IEEEproof}
	The geodesic associated with the distance metric $\dmxtl$ either passes through the position of a landmark or it does not. Assuming that it does not pass through one of the landmarks that the distance metric will be the same as $\dmxt$ since we are not using the identification in $\M_T^L$. If it passes through a landmark then the distance will be equal to $\dmxtl(\e_i,\mathcal{S}_k)+\dmxtl(\mathcal{S}_k,\mathcal{S}_{k'})+\dmxtl(\mathcal{S}_{k'},\e_j)$. However, we know that $\dmxtl(\mathcal{S}_k,\mathcal{S}_{k'}) = \dm(\mathcal{S}_k,\mathcal{S}_{k'})$ by Lemma \ref{lem:dmxtldm}, and we can similarly show that $\dmxtl(\e_i,\mathcal{S}_k) = \dm(\e_i,\mathcal{S}_k)$ and $\dmxtl(\mathcal{S}_{k'},\e_j) = \dm(\mathcal{S}_{k'},\e_j)$. Hence, taking the minimum of both possible scenarios we get the desired result.
\end{IEEEproof}
Finally, combining the result of Lemma \ref{lem:dmxtldm2} with Theorem \ref{thm:1}, we obtain the desired bound on landmark communities.
\begin{theorem}\label{thm:2}
	For any  pair of landmark communities $ \C_k, \C_{k'}$, provided that Assumptions 1 and 2 hold, and  $ 4\delta_e < \varepsilon $ then
	\begin{equation}
	\dm(\C_k, \C_{k'})   \leq   \dgel(\C_k, \C_{k'})   \leq (1+\lambda_3) \cdot \dm(\C_k, \C_{k'}),   
	\end{equation}\label{eq:thm}
	where $\lambda_3 = \tau/\varepsilon - 1 + 4\delta_e \tau / \varepsilon^2$.
\end{theorem}
\begin{IEEEproof}
	The inequality $\dmxtl(\E_k, \E_{k'})   \leq   \dgel(\E_k, \E_{k'})$ can be shown by following the same arguments as in Lemma \ref{lem:1}. Hence, $\dmxt(\E_k, \E_{k'})   \leq   \dgel(\E_k, \E_{k'})$ follows by Lemma \ref{lem:dmxtldm}. Similarly, the second inequality follows by using the argument in Lemma \ref{lem:2} and the results in Lemma \ref{lem:dmxtldm}.
\end{IEEEproof}

Note that we can also generalize the results of this theorem to include bounds on the convergence of distances between any arbitrary events 
by using Lemma \ref{lem:dmxtldm2}. However, this result does not provide convergence to $\dm$ for an arbitrary pair of events, and will depend on an additional constraint on the density of the configuration of landmarks in $\M$. 
\begin{assumption}[Landmark Density Condition] \label{asum:3}
	For any landmark $l_k$, there exists another landmark $l_{k'}$ such that $  d_\M (\C_k, \C_k') \leq \delta_l $. Moreover, for any two distinct landmark communities $d_\M ({\C_k}, \C_{k'}) > 0$. 
\end{assumption}
\begin{cor}\label{cor:1}
	Given two events $\e_i$ and $\e_j$, provided that Assumptions 1-3 hold, and  $ 4\delta_e < \varepsilon $ then
	\beq \begin{array}{l}
		\dm(\e_i,\e_j) \leq \dgel(\e_i,\e_j) \leq (1 + \lambda_3) \cdot \left(\dm(\e_i,\e_j)  + 2 \, \delta_l\right). 
	\end{array}
	\eeq
\end{cor}
\begin{IEEEproof}
	The left hand side inequality can be obtained by following the same arguments as in Lemma \ref{lem:1} to  show that $\dmxtl(\e_i, \e_{j})   \leq   \dgel(\e_i, \e_{j})$, and using the fact that $\dm (\e_i, \e_j) \leq \dmxtl(\e_i, \e_{j})$ from the argument in Lemma \ref{lem:dmxtldm}. 
For the second inequality, following the same argument as in Theorem \ref{thm:1} and Lemma  \ref{lem:dmxtldm2} we have  $\dge(\e_i, \e_j)   \leq (1+\lambda_3) \cdot \dmxtl(\e_i,\e_j)$ and $ \dmxtl(\e_i,\e_j) \leq  \inf_{l_k,l_{k'} \in L} \{ \dm(\e_i,\mathcal{S}_k)+\dm(\mathcal{S}_k,\mathcal{S}_{k'})+\dm(\mathcal{S}_{k'},\e_j) \}$. In particular, if we let $k_1$ and $k_2$ be the landmarks closest to $\mathcal{S}_k$ and $\mathcal{S}_{k'}$ respectively, then $  \dmxtl(\e_i,\e_j) \leq \delta_l/2 + \dm(\mathcal{S}_{k_1},\mathcal{S}_{k_2}) + \delta_l/2 \leq \delta_l + \dm(\mathcal{S}_{k_1},\e_i) + \dm(\e_i,\e_j) + \dm(\e_j,\mathcal{S}_{k_2}) \leq 2 \delta_l + \dm(\e_i,\e_j)$. 
\end{IEEEproof}

Theorem \ref{thm:2} asserts that given enough number of encounter samples from the environment and graph connectivity condition, if the CRW motion of the biobots is close enough to straight line segments for small enough neighborhoods, the proposed encounter metric asymptotically recovers the geodesic distances between landmarks. 
In particular, for a fixed size and shape of the environment, and the speed of agents, the encounter sampling density $\delta_e$ is inversely proportional to $|\E|$, which is proportional to the number of moving agents $N_m$. Hence, with increased number of moving agents, $\delta_e$ will eventually converge to $0$ and $\lambda_1$ can be chosen arbitrary small. $\tau$, on the other hand, is dependent on CRW parameters, the existence of obstacles in the space, and the encounter sampling as well. 

Consider a local subdomain (patch) of $\M$, where there are no obstacles within.
For this patch, $\tau$ is mainly determined by the deviation of the trajectories agents traverse between their encounters from straight line segments. If the agents move on straight lines and change direction only if they detect other agents or obstacles, then: for any two neighboring encounters $\de (\e_i, \e_{j}) = \dmxt (\e_i, \e_j)$, so $\delta_e \leq \varepsilon$ and $\lambda_2 \rightarrow 0$. This depends on the probability of each line segment in CRW model being larger than encounter sampling density. 
Considering the CRW model of (\ref{eq:crw}), $p[\li \geq \delta_e]  = e^{\delta_e/\li^*}$, which will converge to $1$ if $ \delta_e/\li^* \rightarrow 0$. Hence, for a fixed $\li^*$, in an obstacle free space if $\delta_e \rightarrow 0$, then $\lambda_2 \rightarrow 0$. Convergence of $\lambda_2$ also can realized by increasing $\li^*$ for a fixed sampling density, although smaller $\li^*$ will result in better dispersion performance, which is representing a trade-off between exploration and mapping performance. 
When there exist obstacles, the change of direction due to obstacle avoidance introduces an extra source of error, which can also be decreased by having enough number samples close to the boundaries of obstacles. 


\section{Topological Inference}
\label{sec:topInf}
So far, we have constructed point clouds  for geometric estimation of the domain $\M$. However, such models are not precise and is subject to errors due to uncertainty on the distances, which may lead to deformations in the shape of the space. In order to address this issue, we exploit persistent homology, that is stable to bounded perturbations of the dissimilarity measure \cite{Cohen-Steiner2007}, along with a statistical classification approach in order to infer topological features of $\M$ from coordinate-free encounter point clouds. 
 
Consider again the point cloud  ${\E}$ with the encounter metric $\dgel$. We construct filtrations of  
Vietoris-Rips simplicial complexes $\mathcal{R}(\E, \epsilon)$, denoted as \textit{encounter complexes}, and compute its persistent homology to extract topological features in terms of persistence diagrams. An example of a simplicial complex constructed over $\E$ is shown in Fig. \ref{fig:top-map}(d). 
Similarly, one can calculate the persistence over the filtrations of simplicial complexes  $\mathcal{R}(\C, \epsilon)$.
%
%
%
%
We restrict our computations to the dimensions $ 0 $ and $ 1 $ of persistence (corresponding to connected components and holes/obstacles in the point cloud) as the higher dimension are not applicable in our experiments. 
This process is, however, preceded by a subsampling step for cleaning up the data. 

\textit{{Subsampling and Outlier Removal}} \label{subs}
Extracting topological information from the whole point cloud would be computationally expensive due to the large number of events that are created over time, and it can be prone to error due to outliers in  the estimated metric; hence using a subsampling algorithm would be beneficial. 
A common subsampling method in persistent homology which preserves topology is the \textit{maxmin} algorithm, but it is very sensitive to outliers. Hence, we employ a modified version of maxmin by adding  
 a pre-processing layer for outlier removal using a \textit{K-nearest neighbor} (KNN) density estimation \cite{Carl07}. 
For a node $ \e_i $ in graph $\G$, let $ d_k(\e_i) $ denote the distance between $\e_i$
and it's k-nearest neighbor, and $ \overline{d}_k(\e_i) $ be the average distance  to its k-nearest neighbors (inversely proportional to the local density of the point cloud around $ \e_i $). Consider the density of all average  distances over $\E$  as $\rho_k ({\E})$ and select the threshold $\tau_{q,k}$ to be $q$-quantile of $\rho_k ({\E})$. 
Then we select a subset $\V \subset \E$ of points $ v_i $ such that $d_k(v_i) <\tau_{q,k}, \, \forall v_i \in \V $, which will remove outliers of ${\E}$  to a large extent. Finally, we apply the maxmin algorithm on the pre-filtered point cloud $\V$ for further subsampling. 
The first sample $x_1$ is chosen  randomly from the point cloud  $\V$.  Iteratively, if $X=\{x_1, ..., x_{k-1}\}$ is the set of previously selected  samples, find the next sample as a point $x_{k} \in \V $ s.t. maximizes $d(x_k,X)$.
 \subsection{Topological Stability Analysis}
 Here we will analyze topological stability of our metric estimation, and show that our geometrically estimated point cloud will provide estimations of features of environment that are topologically stable. 
 Consider the metric spaces $(\M, \dm)$, $(\E, \dgel),$ and $ (\C, \dgel)$,  
 and let $\pdm, \pdp, \pdl$ represent the corresponding persistence diagrams obtained from filtrations of simplicial complexes $\mathcal{R}(\M, \epsilon)$, $\mathcal{R}(\E, \epsilon)$, $\mathcal{R}(\C, \epsilon)$, respectively.
 We will exploit the bounds obtained in section \ref{sec:metric-analysis} in order to show stability of persistence diagrams over  $\E$ and $\Cs$. 
 
 \begin{theorem}[Stability of persistence]  \label{thm:3} Under assumptions and conditions of Theorem (\ref{thm:1}) and Theorem (\ref{thm:2}), the proposed encounter metric is topologically stable i.e. the persistence diagrams 
 	$\pdp$ and $\pdl$ 
 are not much different form   	$\pdm$ in terms of the bottleneck distance: 
 	\begin{eqnarray}
 	\db(\pdm, \pdl)   & \leq &   \lambda_3  \cdot \delta_l + \delta_l,\label{eq:db-1} \\
 	\db(\pdm, \pdp) & \leq &   (1 + \lambda_3) \cdot 2 \delta_l +  (1 + \lambda_3) \cdot \delta_e.\label{eq:db-2}
 	\end{eqnarray}
 \end{theorem}

 The bottleneck distances between the persistence diagrams 
 in (\ref{eq:db-1}) and (\ref{eq:db-2}) are bounded by the errors that correspond to the summation of  \textit{geometric estimation error} and \textit{sampling error}. 
 In  (\ref{eq:db-1}) the term $\lambda_3 \delta_l$ corresponds to the metric estimation error between landmarks, which will vanish if $\lambda_3 \rightarrow 0$ as discussed in the previous section. The term $\delta_l$, on the other hand, is purely the density of landmarks in the space, and determines the topological estimation error using only landmarks as the samples of the $\M$ if we had the true metric over the space $\C$. Likewise in (\ref{eq:db-2}), the first term corresponds to the metric estimation error between all encounters which will vanish if both $\delta_e, \delta_l \rightarrow 0$. The term $(1 + \lambda_3)\delta_e$ corresponds to the topological perturbation using the sampled set of encounters which will converge to zero if $\delta_e \rightarrow 0$. 
 
 \subsection{Topological Classification }
Persistent diagrams provide a means for extracting topological features out of the point cloud, in terms of 
birth and death of features as the value of $\varepsilon$ varies, encoded in persistence intervals. 
 However, one further needs to  classify persistent features from noisy detections that emerge due to subsampling and  metric estimation errors.
%
 For a dense enough sampled data, however, one can find a range of values for which the homology of simplicial complexes is equal to the one for $\M$ \cite{Bal13}.  
 Here, we propose 
 a robust and scale-invariant 
 classification algorithm for persistence intervals by learning such thresholds based on a non-parametric density estimation of noise in order to separate persistent features form noisy ones and infer the correct Betti numbers of the underlying space.
 %
 
 For our statistical analysis, we consider a collection of sampled point clouds $\{ \E^{(p)} \}_{p=1}^N$ of a topological space $\M$. For each $\E^{(p)}$, we can extract from its persistence diagram a collection of $n$-th Betti number persistent interval lengths $\{ r_n^{(p),k} \}_k,$ where $r_n^{(p),k} = |d_n^{(p),k} - b_n^{(p),k}|$. Denote the $q_n$ quantile for this collection by $r_{q_n}^{(p)}$, and define the $n$-th estimated Betti number for the $p$-th point cloud as
 \beq
 \hat{\beta}_{n}^{(p)}(\theta_n) = \sum_k \mathds{1}_{\mathbb{R}^+}    \left( \frac{r^{(p),k}_n}{r_{q_n}^{(p)} + \Delta_n } - \tau_n   \right),
 \eeq
 where $\theta_n = (q_n,\Delta_n,\tau_n)$ is the estimation parameter vector, and  $\mathds{1}_{\mathbb{R}^+} (.)  $ is the indicator function of ${\mathbb{R}^+}$. This function specifies that only those persistent intervals for which the normalized value of $r_n^{(p),k}$ is greater than $\tau_n$ are considered as ``true'' persistent features. Furthermore, define the $n$-th Betti number error as
 \beq
 e_n^{(p)}(\theta_n) = \left| \hat{\beta}_{n}^{(p)}(\theta_n) - \beta_n\right|,
 \eeq
 where $\beta_n$ is the $n$-th Betti number of $M$. Hence, the average error for the collection of sampled points can be expressed as
 \beq
 e_n(\theta_n) = \frac{1}{N}\sum_{p=1}^N e_n^{(p)}(\theta_n).
 \eeq
 
 In order to perform classification with this metric, we generate a set of samples from the same topological space and train our classifier by finding the value of $\theta_n$ that minimizes the error function $e_n$. In order to make the cost function smooth for optimization purposes, one can replace the indicator function $ \mathds{1}(x) $
 with a sigmoid function $ \sigma_{\alpha}(x) = \frac{1}{1+e^{-\alpha x}}$, where $\alpha$ is a scaling parameter. We denote by $\theta_n^*$ the optimal parameter value found using brute force  optimization. Finally, we can compute the error $e_n(\theta_n^*)$ for a new collection of sampled data for testing purposes.

\section{Numerical Analysis Results} 
\label{sec:SimulationResults}
\begin{figure*}[tbph]
	\begin{center}
				\includegraphics[width=0.95\linewidth]{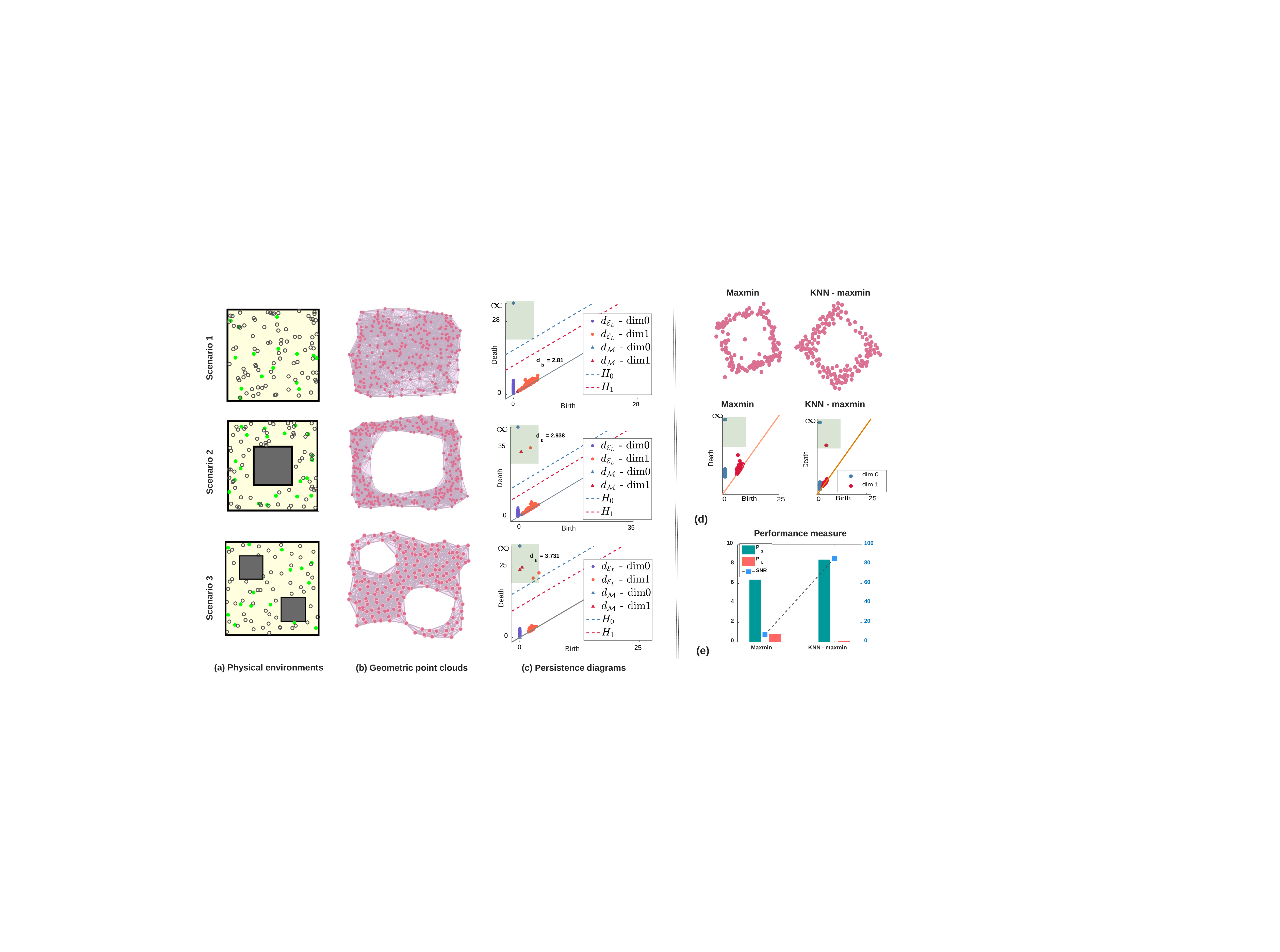} 
	\end{center}
	\caption{ \small{Simulation results for example scenarios (1) - (3): (a) a snapshot of the physical environments with moving agents (gray circles) and landmarks (green circles), (b) simplicial complexes built over learned geometric point cloud, (c)  persistence diagrams $\pdp$ $\pdm$, their bottleneck distances, and topological classification hyperplanes $H_0$ and $H_1$.
			 Subsampling performance: (d) comparison of subsampled point clouds from standard maxmin and KNN-maxmin, and their persistence diagrams, and (e) their SNR performance for 100 independent runs.}}  \label{fig:sim-res}
\end{figure*}
In this section, we present numerical simulation experiments to demonstrate the validity and performance of our approach. We have considered two-dimensional environments with basic geometric shapes for simplicity. However, the approach makes no use of these assumption and is directly applicable to more unstructured environments. 
Additionally, videos of the simulations for more complex environments as well as experiments with Wolfbots are available in the project website \footnote{\url{https://research.ece.ncsu.edu/aros/project/}}.
The software is composed of a {biobotic swarm simulator} and a {GTP-mapping} package. The swarm simulator (in C++) reproduces the behavior of biobots by combining natural as well as control motion patterns. The mapping package (in C++ and Matlab) performs geometric estimation and topological inference.
In the swarm simulator, CRW parameters are set to be $ C=1, \li^*  = 0.30\, m,  v_m = 0.1 m/s $. The change of direction is isotropic everywhere except the boundaries of the environment. The detection radius of each agent for encounter and boundary recognition is adopted to be $ r_d = 1cm$. 
For computation of persistence intervals and persistent diagrams, we exploit
Dionysus C++ library  \cite{Dion}.

\subsection{Example Scenarios}

Here we illustrate our methodology presenting the results for three different example scenarios as shown in Fig. \ref{fig:sim-res}. We consider square shaped physical environments with dimensions $50m \times 50 m$ and three scenarios as (1) no obstacles, (2) one obstacle in the middle, and (3) two obstacles.   
A total number of 150 agents are considered to start their CRW motion from random initial configuration among which $N_l = 20$ are selected as landmarks after the initial dispersion and are commanded to switch to S mode, while the others ($N_m = 130$) keep moving as CRW motion, and exploring the area for 200 seconds.  Fig~\ref{fig:sim-res}(a) sketches  snapshots of the  environments together with moving agents (gray circles) at $ t=0 $ and the set of landmark nodes (shown as pink dots).

We constructed estimated geometric point clouds and persistence diagrams for each scenario. Simplicial representations of each point cloud in the form of Rips complexes $\mathcal{R}(\E, \epsilon)$, where $\epsilon  = 0.5 \, 
\max({\dgel})$, are shown in Fig.~\ref{fig:sim-res}(b). The obstacles in the physical environments are represented as holes in these representations.
\begin{figure*}[tbph]
	\begin{center}
		\includegraphics[width=0.95\linewidth]{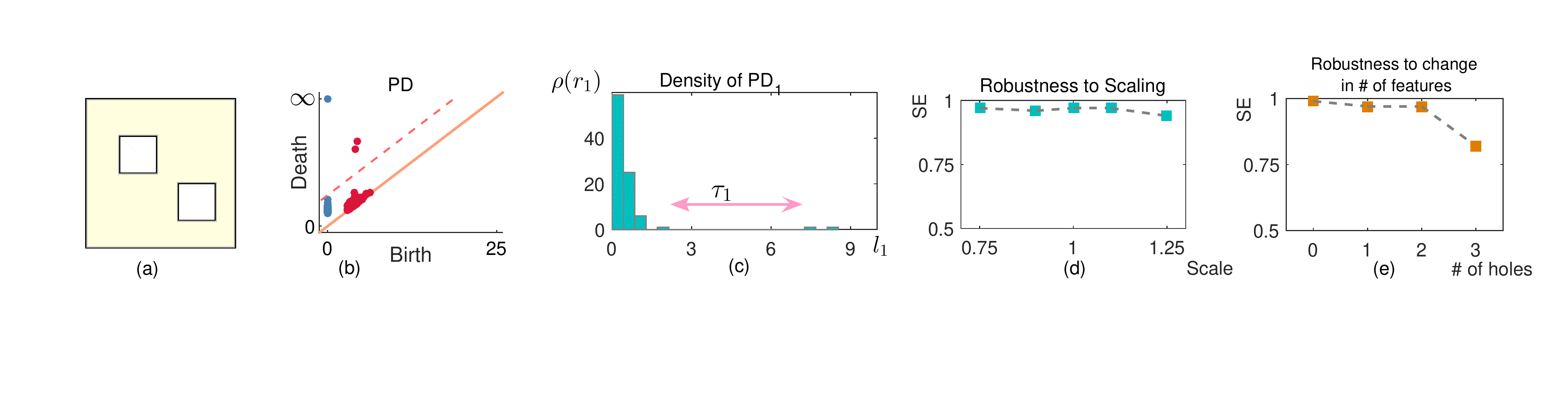} 
	\end{center}
	\caption{ \small{
			Robust Classification of persistence intervals: (a) A 2-hole environment, (b) the extracted persistence diagram, (c) density of persistence intervals with the appropriate threshold,  (d) robustness of classification algorithm to scaling,  and (e)  to change in  the number of features.}}  \label{fig:clust}
\end{figure*}
\begin{figure}[b!]
	\centering
	\includegraphics[width=0.97\linewidth]{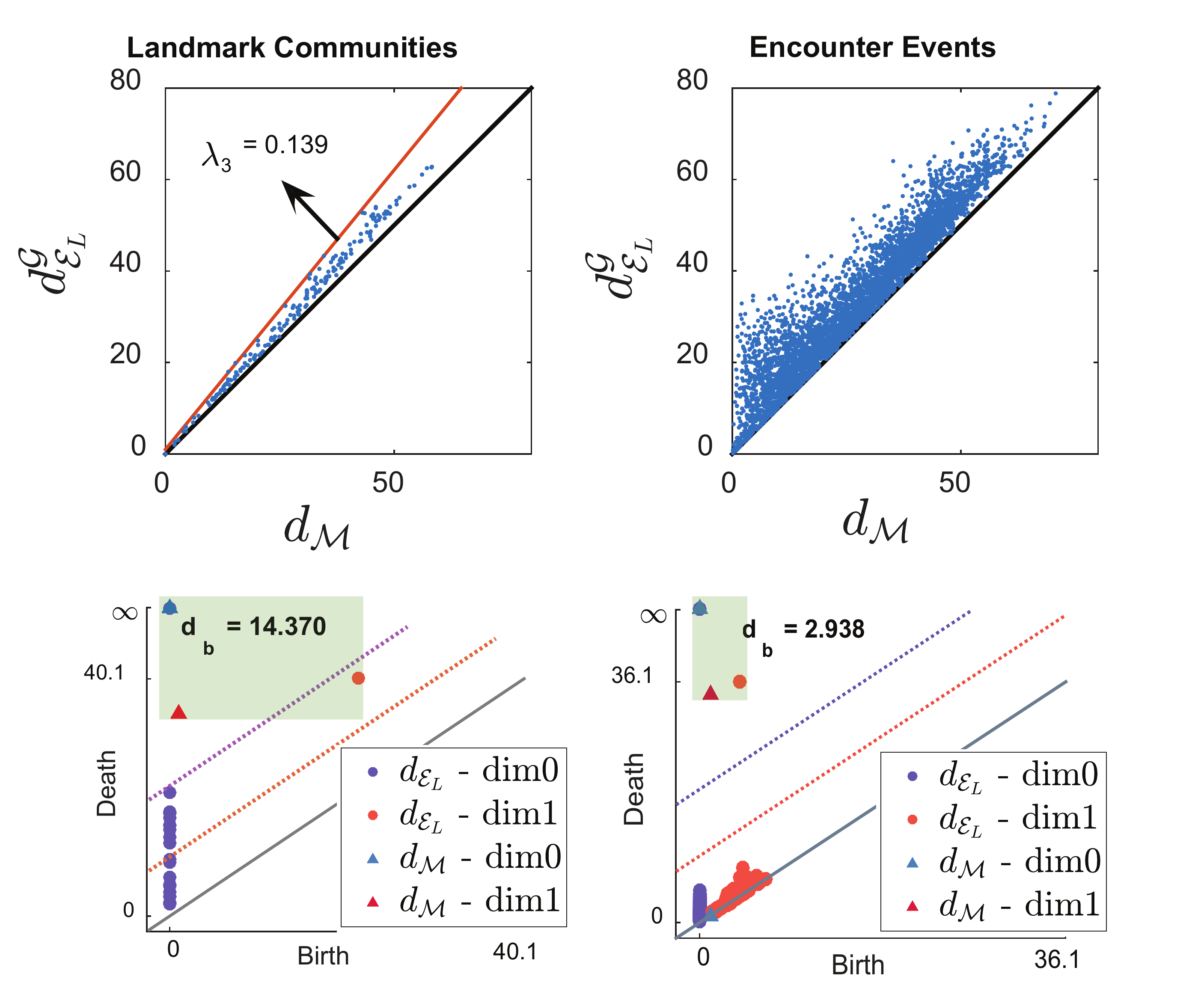}
	\caption{ \small{ Estimated geodesic distances $\dgel$ vs. $\dm$ and the corresponding $\pd$'s over the set of landmark communities (left), and the set of encounters (right) for the environment in scenario 2.}} 
	\label{fig:dgm-de}
\end{figure}

The  persistent diagrams for the filtrations over the point clouds are sketched on top of the true persistence diagrams of the spaces in Fig. (\ref{fig:sim-res}). 
On each persistence diagram, 0-dimensional and 1-dimensional topological features (connected components and holes)  are plotted as ``blue'' and ``red'' points, respectively, and the class hyperplanes $H_0$ and $H_1$ separating topological features from noise are drawn as dashed lines. 
The   classification thresholds are obtained after 50 independent runs as $\tau_0  = 25, \tau_1 = 30$. The features classified as signals are marked with green rectangular patches. 
On each diagram, we calculated the bottleneck distance between the persistence intervals obtained over the filtration of the true manifold distances and the encounter metric $\db(\pdm, \pdp)$. The maximum topological estimation errors are in the order of $4.2\%$ of the largest pairwise distance value in the environments. 
  It can be observed that the point clouds capture the correct topological features of the domains, which can  be verified by the features classified as signals  in the persistence diagrams. Moreover, the estimated point clouds also contain some geometric information, which could be exploited to improve the mapping accuracy for approaches that make use of additional sensing and localization information.   

In order to explore the effect of point cloud subsampling on the topological inference, we further studied the signal to noise ratio (SNR) performance over $ N = 100$ independent runs for the one hole case, comparing the standard maxmin subsampling with the proposed KNN pre-filtered maxmin approach. 
Fig.~\ref{fig:sim-res}(e)-(f) shows the subsampled point clouds consisting of 150 points along with the corresponding persistence diagrams. KNN parameters are selected as  $ q=0.9 $ and $ k=10$.

The signal to noise ratio is calculated as $ SNR = { P_S }/{P_N}$ where $P_S$ and $P_N$ represent the average of persistence signal power  $P_S^{(p)}$ and noise power $P_N^{(p)}$ over $N$  runs, respectively. We define persistence signal power for the $p$-th data set as 
$  P_S^{(p)} =~\sum_{\mathpzc{S}^{p}} |{r_1^{(p),k}}|^2 /|\mathpzc{S}^{p}|$, where $\mathpzc{S}^{p}$ is  the signal set for the $p$-th run  consisting of all persistence intervals of dimension 1 classified as signals, and ${r_1^{(p),k}}$ represents the corresponding lengths of such intervals. Similarly, $P_N^{(p)} =~\sum_{\mathpzc{N}^{p}} |{r_1^{(p),k}}|^2 /|\mathpzc{N}^{p}|$, where $\mathpzc{N}^{p}$ denotes the $p$-th noise set. 
An investigation of the plots in this figure clarifies how the pre-filtering step cleans up the data from outliers that fill out the topological holes, and  improves the SNR to a great extent.
\subsection{Classification and Robustness} 
Additionally, we assessed the performance of our topological classification in terms of robustness to scaling and number of holes in the space. We used a training data set consisting of 100 variations of the space shown in figure \ref{fig:clust}(a) with random placements of the two square holes in the space.  Persistence diagram and the corresponding density of interval lengths for one of the environments is shown in figure \ref{fig:clust}(b) and (c), respectively. 
The optimization process resulted in a minimum of  $ \mathcal{F} (\Theta^*_1)  = 0.03$ with  $\Theta^*_1 = (0.5 , 0.7,  37)$.  This result is not surprising as the $0.5$-quantile of the density, which is the median of the dataset, is known to be robust to outliers. 
We evaluated $\mathcal{F} (\Theta^*_1)$ for a variety of  test sets, each containing  100 point clouds with random feature placements, with each test set differing in either scale of the environment and features or the number of features (holes). 

To investigate the performance, we used  the sensitivity measure for classification, $ SE  = TP/ (TP+FN)$, where $ TP$ and $ FN $ denote the number of true positives and false negatives, respectively. The sensitivity performance for 
these scenarios is summarized in Fig. \ref{fig:clust} (d) and (e) for scaled environments, and environments with different numbers of holes but the same scale, respectively. Note that we have evaluated this measure for the whole mapping algorithm and not just for the classification part, which can justify the degradation in the performance for the 3-hole case, as we observed that in $15\%$ of the cases, the third hole could not be retrieved from the point cloud data. The reason is that adding more topological holes in a fixed space will degrade the encounter graph connectivity, which which in turn deteriorates the geometric estimation accuracy and  topological connectivity. 
\subsection{Convergence Analysis}
We performed a set of experiments in order to evaluate the geometric estimation error and topological stability bounds derived in sections \ref{sec:topInf} and \ref{sec:Metric}. 
\begin{figure}[b!]
	\centering
	\includegraphics[width=0.93\linewidth]{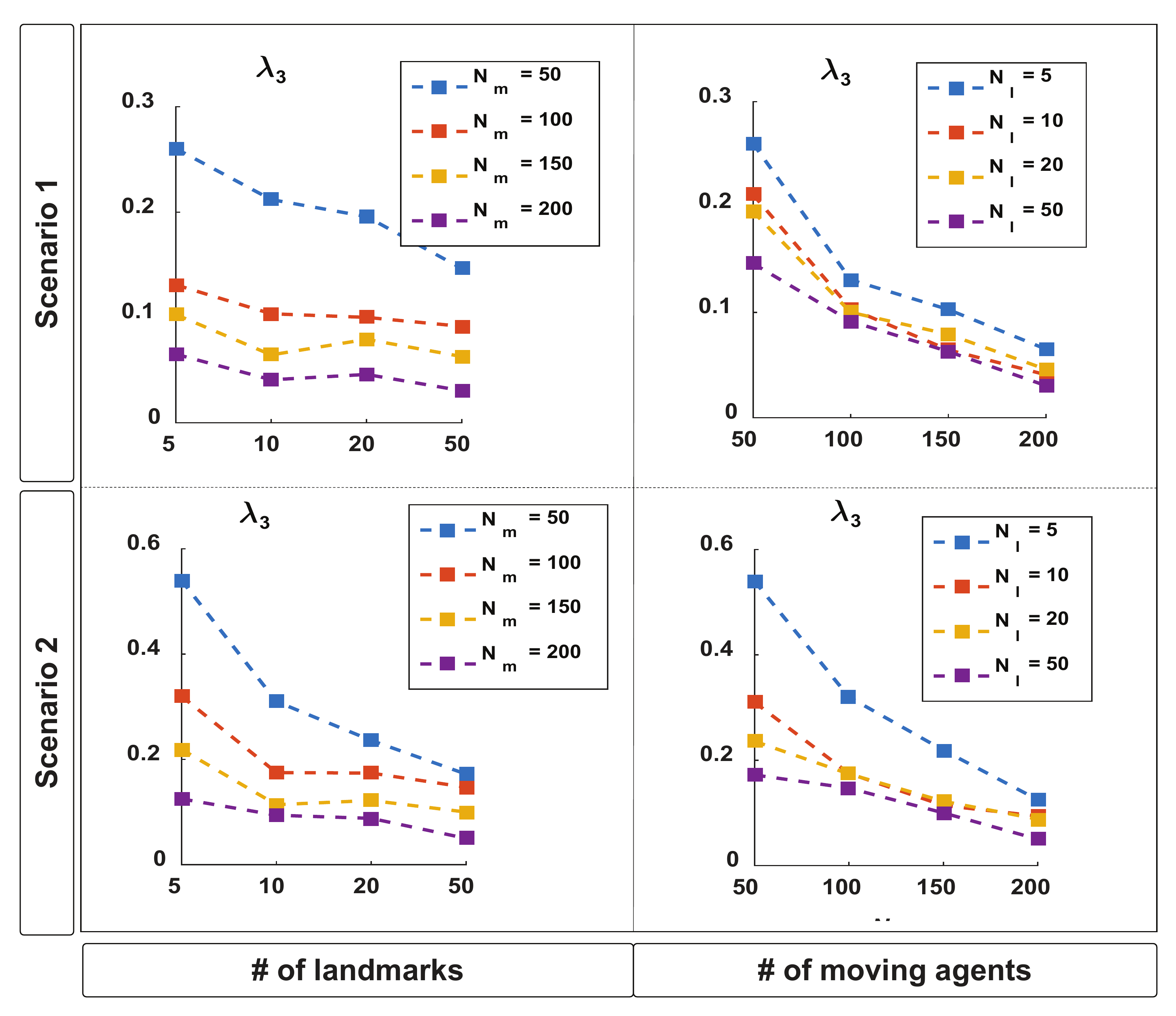}
	\caption{ \small{ Variation of the geometric estimation convergence parameter $\lambda_3$ with $N_l$ and $N_m$. Increasing $N_l$ for different fixed values of $N_m$ (left), and increasing $N_m$ with fixed values of $N_l$ (right). }} 
	\label{fig:bounds-lambda}
\end{figure}
Fig. \ref{fig:dgm-de} sketches the estimated geodesic distances versus the manifold distances over the sets $\Ls$ and $\Ps$  for a single run with $N_l = 20$ landmarks and $N_m = 100$ moving agents, using the same parameters described as in example scenarios.
 One can observe how closely $\dgel$ estimates geodesic distances over the set of landmarks, while it is prone to more error for pairs of non-landmark events, which is consistent with the results in Theorem \ref{thm:2}. However, looking at the bottleneck distance $\db(\pdm, \pdp) $
 in the right plot, we realize that the robustness of persistent homology to perturbations compensates for the relative large  geometric errors yielding to a bottleneck distance of only 2.938 in topological estimation. The relatively larger topological error $ \db(\pdm, \pdl)$ comes from the fact that although the metric $\dgel$ among the landmarks is more precisely estimated, the density of the landmarks (inversely proportional to $\delta_l = 7.27$) is not sufficient to capture the topological information with a high precision if the simplicial complexes are constructed on a subset of $\M$ consisting of only $\Pi_\M(\C)$.
 Note that the density of $\Pi_\M(\C)$  is in fact equal to the density of position of the landmarks themselves as they do not move over time, whereas for moving agents, densities of their encounter locations $\Pi_\M(\E)$  in $\M$ increases over time.   
  Hence although the metric $\dgel$ learned over $\C$ is estimated more precisely, topological inference is better to be carried out over the set of $\E$. The larger density of samples $\Pi_\M(\E)$ with topological stability of persistence homology will compensate for  the larger geometric estimation  errors of $\dgel$ over $\E$. 
  
\begin{figure}[t!]
	\centering
	\includegraphics[width=0.99\linewidth]{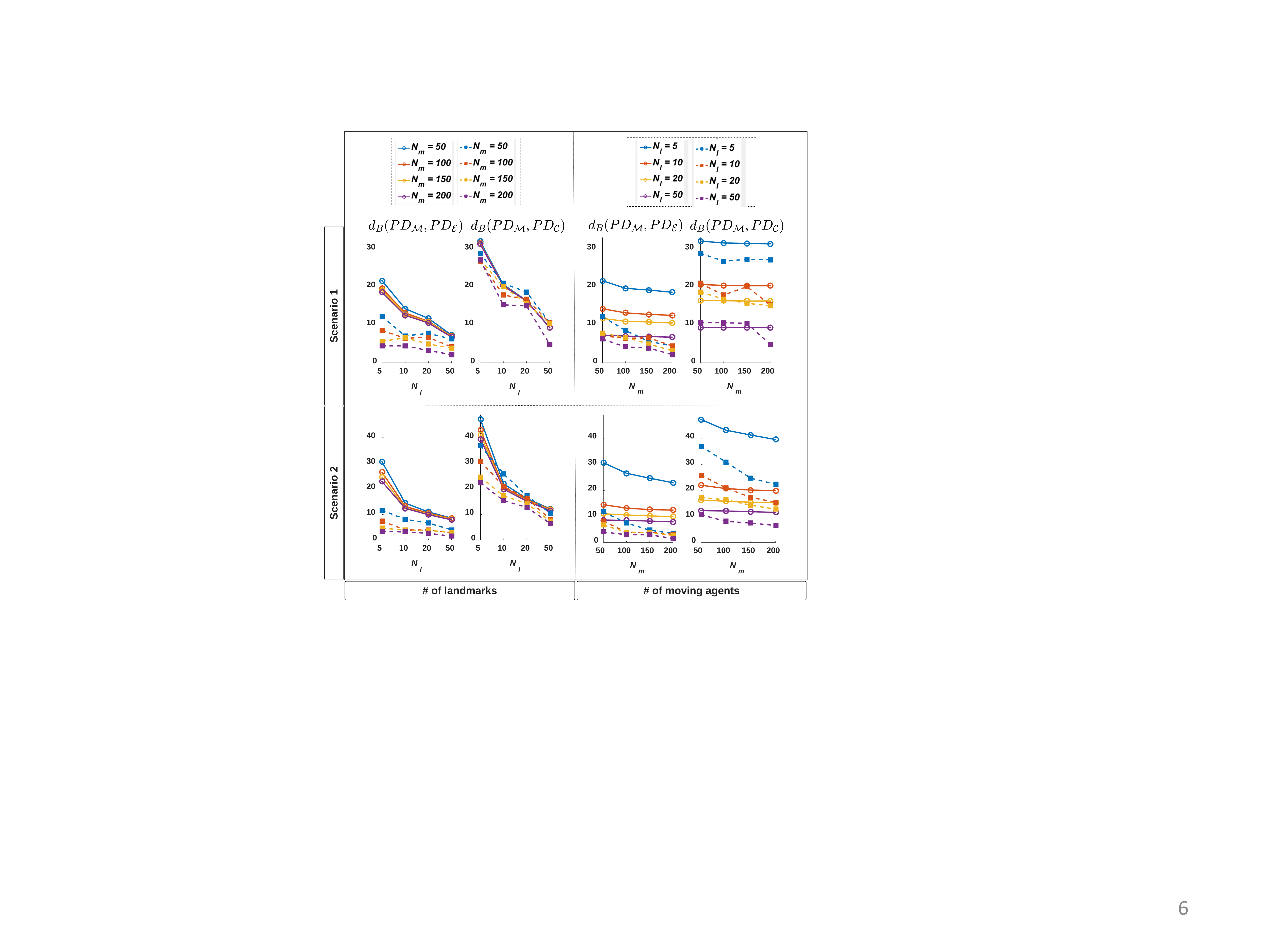}
	\caption{ \small{Variation of the bottleneck distances $\db(\pdm, \pdp)$ and $\db(\pdm, \pdl)$ (dashed lines) and estimated stability bounds (solid lines) with $N_l$ and $N_m$.  Increasing $N_l$ for fixed values of $N_m$ (left),  and increasing $N_m$ with fixed values of $N_l$ (right).}} 
	\label{fig:bounds-db}
\end{figure}

 This conveys a trade-off between learning the geometry and topology in terms the densities of landmarks and encounter samples. To better illustrate this, 
  we studied the variations of the geometric convergence parameter $\lambda_3$  as well as the bottleneck distances $\db(\pdm, \pdp) $ and $ \db(\pdm, \pdl)$
  in terms of  $N_m$ and $N_l$ for scenarios 1 and 2.   
Particularly, we considered two case studies: (1) increasing $N_m$ for fixed values of $N_l$, and (2) increasing $N_l$ for fixed $N_m$. For each case, we considered 20 independent runs,  and four different values for the controlled parameter. The trends for independent parameters  $\lambda_3$ and bottleneck distances are shown in the diagrams plotted in Fig. \ref{fig:bounds-lambda}  and Fig.~\ref{fig:bounds-db}, respectively. 

A quick overview of these diagrams verifies that the metric estimation errors and bottleneck distances will both generally decrease with increasing values of $N_l$ and $N_m$.
Looking more carefully at  Fig.~\ref{fig:bounds-lambda} one can easily observe the monotonously decreasing pattern of the values of $\lambda_3$  as $N_m$ increases, and the higher sensitivity of $\lambda_3$ with respect to $N_m$ than $N_l$. This conveys the fact that although landmarks are essential for precise geometric estimation, increasing their number would not improve the accuracy of $\dgel$ metric over $\C$ as much as increasing $N_m$ will do. However, increasing $N_l$ will improve the accuracy of $\dgel$ over $\E$
to a greater extent as a consequence of Theorem \ref{thm:2} and Lemma \ref{lem:dmxtldm}. 
On the other hand, from Fig.~\ref{fig:bounds-db}, one can observe that $N_l$ has a greater impact on topological estimation over landmarks than $N_m$. This follows the same discussion mentioned for bottleneck distances in Fig.~\ref{fig:dgm-de}, as generally $\db(\pdm,\pdl)$ is dominated by $\delta_l$. Moreover, looking into diagrams for $\db(\pdm,\pdp)$ in Fig.~\ref{fig:bounds-db}, one notices that the topological stability bounds in this case are dominated by $N_l$ rather than $N_m$, although the experimental trends for the two cases do not vary that much. Furthermore, it is evident that the bottleneck distance $\db(\pdm,\pdp)$ are  in general smaller than $\db(\pdm,\pdl)$ due to the same argument as mentioned for Fig.~\ref{fig:dgm-de}. 
Hence, having a fixed number of agents at hand,  a reasonable strategy would be to select a small percentage (10 - 20 $\%$) of them as landmarks, in order to improve geometric estimation over the set of encounters, and infer topological structures over the set $\E$. 
Then the ratios of $N_m$ and $N_l$ can further be adjusted based on the needs of the problem and the priorities of geometric or topological learning. Note that in our analysis above, we discussed $N_m$ as the defining parameter for density of events, while other parameters are fixed. However, one can also increase the running time of the exploration stage or the speed of the agents depending on the problem constraints in order to acquire more encounter samples so as to increase the accuracy with smaller number of agents.

\section{ Conclusion and Future Work} \label{sec:conclusion}
We presented a framework for geometric-topological learning of unknown environments using local interactions amongst the agents in biobotic networks. Geodesic metric estimation and topological stability convergence analysis was provided, and it was shown that under dense sampling of the environment, connectivity of the network encounter graph, and existence of enough landmarks, one can learn the intrinsic geometry of the environment with arbitrary small errors, extract topological features of the space with bounded error, and  infer the true homology of the space with the proposed classifier. 
Our future work will include the results of experiments for implementation of the proposed approach on Madagascar hissing biobots in order to realize exploration and mapping with cyborg insect networks for search and rescue operations.

\ifCLASSOPTIONcaptionsoff
  \newpage
\fi



%
%
%

\bibliographystyle{myIEEEtran}
\bibliography{Alireza-bib}
\end{document}